\pdfoutput=1
\documentclass[10pt,twocolumn,letterpaper]{article}

\usepackage[letterpaper,textwidth=6.875in,textheight=9in,columnsep=0.3125in,centering]{geometry}
\usepackage[T1]{fontenc}
\usepackage[utf8]{inputenc}
\usepackage{amsmath}
\usepackage{newtxtext}
\usepackage{newtxmath}
\usepackage{microtype}
\usepackage{xspace}
\usepackage{booktabs}
\usepackage{array}
\usepackage{multirow}
\usepackage{xcolor}
\usepackage{graphicx}
\usepackage{comment}
\usepackage[font=small,labelfont=bf]{caption}
\usepackage{titlesec}
\usepackage[numbers,sort&compress]{natbib}

\titleformat*{\section}{\large\bfseries}
\titleformat*{\subsection}{\normalsize\bfseries}
\titlespacing*{\section}{0pt}{1.6ex plus 0.5ex minus 0.2ex}{0.8ex plus 0.2ex}
\titlespacing*{\subsection}{0pt}{1.2ex plus 0.4ex minus 0.2ex}{0.6ex plus 0.2ex}

\definecolor{linkblue}{RGB}{0, 80, 160}
\usepackage[colorlinks=true,linkcolor=linkblue,citecolor=linkblue,urlcolor=linkblue]{hyperref}
\usepackage[capitalise]{cleveref}

\def\papersrc{.}
\graphicspath{{figures/pdf_figures/}}

\hypersetup{
  pdftitle={Learning Dynamic Grasping},
  pdfauthor={Nga Kiu K. Ho, Yilin Liu, Hao Xu, Niloy J. Mitra}
}

\title{\Large\bfseries Learning Dynamic Grasping}

\author{%
  Nga Kiu Kelly Ho$^{1}$ \qquad
  Yilin Liu$^{1}$ \qquad
  Hao Xu$^{2,3}$ \qquad
  Niloy J. Mitra$^{1,3}$\\[1ex]
  {\normalsize $^{1}$University College London \qquad
  $^{2}$The Chinese University of Hong Kong \qquad
  $^{3}$Adobe Research}\\[0.5ex]
}
\date{}

\begin{document}

\twocolumn[{%
\renewcommand\twocolumn[1][]{#1}%
\maketitle
\vspace{-4ex}
\begin{center}
  \includegraphics[width=\textwidth]{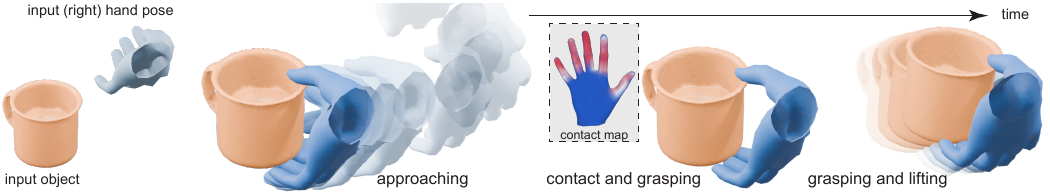}
  \captionof{figure}{Learning dynamic grasping from a single initial hand--object
  configuration. Given only an object and an initial hand pose (left), our
  method generates a complete grasping motion: approach, contact, and lift
  (right). A grasping contact map (see inset) conditions hand poses generated at
  arbitrary normalized times, and object motion is recovered from the hand by
  training-free rigid registration.}
  \label{fig:teaser}
\end{center}
\vspace{2ex}
}]

\begin{abstract}
Synthesizing believable 4D hand--object grasping sequences is important for character animation, extended reality, and embodied agents.
Grasping requires a hand to approach, contact, and lift or manipulate an object coherently.
Existing synthesis methods often model only the final static grasp or rely on additional conditions, such as language, target grasps, or object trajectories, to specify the interaction.
We instead study the underconstrained problem of generating a complete grasping motion from a single initial hand--object state.
Our framework decouples hand-motion generation from object-motion recovery.
Given an object point cloud in the initial wrist frame and the initial parameterized hand configuration, conditional flow matching predicts a sequence-level contact map and a hand pose at each normalized sequence time.
This time-indexed formulation avoids the error accumulation of autoregressive rollout while representing the complete motion.
We then infer an attachment frame and recover rigid-object motion through training-free, contact-guided weighted registration.
We evaluate on held-out DexYCB objects and unseen object geometries reconstructed from in-the-wild videos by Open-CHOIR.
Compared with LatentHOI, our method produces smoother hand and object motion
in every setting, matches or improves on its contact and penetration metrics
in the wild under a shared initialization, and is about 30$\times$ faster per
generated frame. In a 15-participant study, our motions were preferred in
roughly two thirds of (forced-choice pairwise) comparisons. Precise contact closure on novel geometry, however,
remains a key open challenge. Code will be released.
\end{abstract}

\section{Introduction}
\label{sec:intro}

Grasping is a fundamental way in which humans interact with the physical world.
Reproducing it computationally requires more than synthesizing a plausible static hand pose: the hand must approach an object, establish contact, and manipulate it through a temporally coherent motion.
Such \textit{dynamic grasping} is valuable for character animation, virtual and augmented reality, and embodied agents.

Acquiring training data for dynamic interactions, however, remains challenging.
High-quality datasets commonly rely on calibrated multi-camera RGBD rigs or specialized motion-capture systems~\cite{Hampali2020HOnnotate, Taheri2020GRAB, Chao2021DexYCB}.
Monocular reconstruction reduces the (specialized) capture burden, but must resolve coordinated hand and object motion despite severe mutual occlusion and depth ambiguity~\cite{Xu2026OpenCHOIR}.
Consequently, available datasets cover only a limited range of objects and interaction trajectories.
Existing synthesis methods address this ambiguity either by predicting a static grasp endpoint~\cite{Jiang2021GraspTTA, Liu2023ContactGen, Ye2024GHOP} or by introducing additional conditions such as language, target grasps, body motion, or object trajectories~\cite{Christen2022DGrasp, Cha2024Text2HOI, Christen2024DiffH2O, Li2025LatentHOI}.
Generating a complete dynamic grasp from only a single initial hand--object state therefore remains underexplored.

We address this problem through a hand-centric formulation.
We learn how the hand evolves during a grasp while treating the motion of a rigid object as a consequence of \textit{contact} established with the hand.
This decoupling keeps the learned component focused on articulated hand dynamics and recovers object motion geometrically after grasp formation.
We train and evaluate on captured benchmark sequences, and use interactions reconstructed from in-the-wild videos by Open-CHOIR~\cite{Xu2026OpenCHOIR} to test transfer to novel object geometries.
This separates reusable grasping dynamics from object-specific geometry.

Algorithmically, we decompose dynamic grasp generation into two stages.
First, given an initial object point cloud expressed in the initial wrist frame and the corresponding initial MANO~\cite{Romero2017MANO} hand configuration, our conditional flow-matching model~\cite{Lipman2023FlowMatching} predicts how the hand evolves from approach, through contact, to object manipulation.
The framework predicts a sequence-level contact map and a MANO pose at each normalized sequence time, without autoregressive rollout.
Second, given the generated hand trajectory and contact information, we infer an attachment frame and the corresponding hand--object contacts, then rigidly propagate the object through the remainder of the sequence using these correspondences.
Thus, we learn contact and articulated hand motion, then recover object motion geometrically without another learned model.

We train on DexYCB~\cite{Chao2021DexYCB} and evaluate unseen-object generalization on objects  disjoint from training, using both captured and LatentHOI-generated initial hand poses. We further test cross-domain generalization using in-the-wild sequences reconstructed by Open-CHOIR~\cite{Xu2026OpenCHOIR}.
Our evaluation shows that the generated hand trajectories transfer well across this distribution shift, while revealing grasp closure as the primary remaining challenge.
In a 15-participant study, our motions receive 66.25\% preference for naturalness and 64.58\% for apparent grasp plausibility; generation takes 2.8\,ms per frame, versus 85.8\,ms for LatentHOI~\cite{Li2025LatentHOI}.
Further, our training-free object-motion stage propagates unseen rigid objects.
In summary, our contributions are:
(i)~formulating dynamic grasp generation from one initial hand--object state, without observed interaction or text;
(ii)~modeling contact-conditioned hand motion with conditional flow matching;
(iii)~recovering rigid-object motion from inferred contact without further training; and
(iv)~evaluating transfer to unseen DexYCB objects and Open-CHOIR reconstructions, with grasp closure identified as the main challenge.

\section{Related Work}
\label{sec:related}

\paragraph*{Full-body human--object interaction synthesis.}
A substantial body of work models interactions at whole-body and scene scales.
InterDiff~\cite{Xu2023InterDiff} predicts future human and dynamic-object motion from observed interaction history and applies a physics-informed correction during diffusion.
OMOMO~\cite{Li2023OMOMO} reverses this conditioning direction, generating full-body manipulation motion from a supplied object trajectory and using predicted hand positions to promote contact.
Text-conditioned approaches instead generate the human, object, and their contacts jointly: CG-HOI~\cite{Diller2024CGHOI} uses learned contact as guidance, while CHOIS~\cite{Li2024CHOIS} additionally conditions on initial states and sparse object waypoints to support long-horizon interactions in 3D scenes.
Recently, DecHOI~\cite{Jung2026DecHOI} separates trajectory planning from detailed action synthesis and supports responsive replanning in dynamic environments.
These methods coordinate locomotion, balance, and manipulation using past motion, language, object trajectories, waypoints, or scene constraints.
They prioritize body-scale behavior over finger articulation.
We instead focus on grasp formation from a single initial hand--object state.

\paragraph*{Hand animation in computer graphics.}
Early graphics systems used rule-based grasp and hand control~\cite{rijpkema1991knowledge,sanso1994hand}.
Captured examples enabled shape- and task-aware grasp selection~\cite{li2007datadriven}, physics-based controllers~\cite{pollard2005physically,zhao2013robust}, and contact-force-aware interaction transfer~\cite{kry2006interaction}.
Optimization methods used initial contact and kinematic goals~\cite{liu2008synthesis}, an initial grasp and partial object trajectory~\cite{liu2009dextrous}, high-level object goals~\cite{mordatch2012contact}, or captured body and object motion~\cite{ye2012synthesis}.
These works establish contact and dynamics as central to hand animation.
We generate approach--contact--lift motion from an initial hand--object state, without a prescribed goal or object trajectory.

\paragraph*{Static hand grasp synthesis.}
Object-conditioned grasp synthesis commonly predicts a plausible hand configuration at a single instant.
The GRAB dataset introduced GrabNet~\cite{Taheri2020GRAB}, which combines a conditional variational model with refinement to synthesize MANO grasps for unseen object shapes.
GraspTTA~\cite{Jiang2021GraspTTA} further couples hand-pose generation with hand--object contact consistency and adapts the model at test time to out-of-domain objects, while ContactGen~\cite{Liu2023ContactGen} generates an object-centric contact representation before optimizing the hand pose.
More recent diffusion priors improve diversity and geometric generalization: G-HOP~\cite{Ye2024GHOP} jointly models hand configurations and category-level object shape, and S2HGrasp~\cite{Wang2024S2HGrasp} operates on partial, single-view scene point clouds.
Related work on robotic dexterous hands has emphasized physical feasibility at scale. DexGraspNet~\cite{Wang2023DexGraspNet} introduces a differentiable force-closure-based pipeline for synthesizing diverse static grasps, including support for MANO hands, while DexGrasp Anything injects physical constraints during both diffusion training and sampling~\cite{Zhong2025DexGraspAnything}.
Concurrently, DCGrasp~\cite{Akada2026DCGrasp} uses a distance-profile representation to support controllable grasp generation.
These methods generate grasp endpoints rather than complete approach--contact--lift motions.

\paragraph*{Dynamic hand--object motion synthesis.}
Moving beyond grasp endpoints, ManipNet~\cite{Zhang2021ManipNet} synthesizes finger motion from object geometry and manipulation trajectories using a hand--object spatial representation.
D-Grasp~\cite{Christen2022DGrasp} learns a reinforcement-learning policy in simulation for approach, grasp, and transport given a reference grasp and target object pose.
GraspXL~\cite{Zhang2024GraspXL} scales simulation-based control to diverse unseen objects and hand morphologies, with control over grasp region, approach direction, wrist rotation, and hand position.
These methods pursue physically grounded motion but rely on prescribed goals or a simulation environment rather than an initial configuration alone.

Generative methods instead vary the conditioning signal.
Text2HOI~\cite{Cha2024Text2HOI} uses diffusion conditioned on language, object geometry, and generated contact maps, while DiffH2O~\cite{Christen2024DiffH2O} separates grasp acquisition from subsequent manipulation using text and optional target-grasp guidance.
LatentHOI~\cite{Li2025LatentHOI} separates global motion from fine-grained grasping in a latent representation to generalize to unseen objects; its single-hand model and DexYCB evaluation make it our closest experimental baseline (Section~\ref{sec:evaluation}).
HOIGPT~\cite{Huang2025HOIGPT} tokenizes interaction sequences for generation or completion from text, objects, and partial motion, while OpenHOI~\cite{Zhang2025OpenHOI} combines language reasoning with affordance-driven diffusion for open-vocabulary manipulation.
MEgoHand~\cite{Zhou2025MEgoHand} uses flow matching with egocentric RGB, language, and an initial hand pose, whereas HO-Flow~\cite{Chen2026HOFlow} uses latent flow matching to generate joint hand--object motion from text and a canonical object.
Most of these methods generate hand and object motion jointly and use semantic or trajectory-level conditions to resolve the intended action.
In contrast, our learned components predict hand-side contact and articulated hand motion from only a single initial hand--object configuration, without text, target grasps, target object poses, or prescribed trajectories; object motion is estimated subsequently.

\section{Problem Formulation}
\label{sec:problem}

We study the generation of a single-hand grasping sequence from an initial hand--object configuration.
The input consists of an object point cloud with normals,
\begin{equation}
    \mathcal P_0=\{(\mathbf p_i,\mathbf n_i)\}_{i=1}^{N},
\end{equation}
and an initial MANO hand state $\mathbf h_0$~\cite{Romero2017MANO}.
We use the initial wrist frame as the common coordinate system and retain it throughout generation.
The initial wrist orientation and translation are canonicalized to identity and zero, respectively, while the transformed object point cloud encodes the object's position and orientation relative to the hand.
Object positions and normals undergo, under motion, the corresponding rigid transformation and rotation, respectively, and all generated hand orientations and translations are expressed in this fixed frame.

Let $J$ denote the number of articulated MANO joints, let $d_R$ denote the rotation-representation dimension, and define $D_h=(J+1)d_R+3$. We represent hand motion using the vectorized state
\begin{equation}
    \mathbf h=\big[\operatorname{vec}(\mathbf q),\mathbf r\big]
    \in\mathbb R^{D_h},
    \label{eq:hand_state}
\end{equation}
For MANO, $J=15$ and $d_R=6$, so $D_h=99$. Here $\operatorname{vec}(\mathbf q)\in\mathbb R^{(J+1)d_R}$ stacks the continuous 6D rotation representations~\cite{Zhou2019Rotation} of the articulated hand joints and includes the global wrist orientation. The subset $\mathbf q^{\mathrm{art}}\in\mathbb R^{J\times d_R}$ contains the articulated-joint rotations, and $\mathbf r\in\mathbb R^3$ is the wrist translation relative to its initial position.
Each 6D representation is orthonormalized into a valid rotation matrix before evaluating MANO.
We hold the MANO shape coefficients $\boldsymbol\beta$ fixed per sequence and use them only for mesh reconstruction.
The resulting mesh is denoted by $\mathcal M(\mathbf h;\boldsymbol\beta)$, with $F$ faces in a fixed order.
We use a right-hand model and consider one hand at a time.
Left-hand sequences are mirrored to the right hand.

The motion variable $t\in[0,1]$ specifies a normalized position within a finite sequence window, where $t=0$ denotes the initial observation and $t=1$ denotes the final frame.
We define a fixed set of $K$ candidate vertices on the five fingers of the MANO hand mesh. We seek a time-independent, sequence-level contact-distance map $\widehat{\mathbf c}\in[0,1]^K$, shared across all motion times, together with a hand state $\widehat{\mathbf h}(t)$ at each requested motion time $t$.
Each contact-map entry describes the relative proximity of one candidate vertex to the object at the reference contact frame; zero and one denote the closest and farthest candidates within that map, respectively.

After generating the hand poses, a geometric procedure estimates an attachment time $t_{\mathrm{att}}$ and recovers rigid-object motion from that frame onward.
Let $T_k\in\mathrm{SE}(3)$ map the initial object points to their positions at a sampled motion time $t_k$.
The object remains at its initial pose until the selected attachment frame, so $T_k=I$ before and at that frame.
Note that neither the attachment time nor the object transforms are direct outputs of a learned prediction head.

\section{Method}
\label{sec:method}
\begin{figure*}[!t]
    \centering
\includegraphics[width=\linewidth]{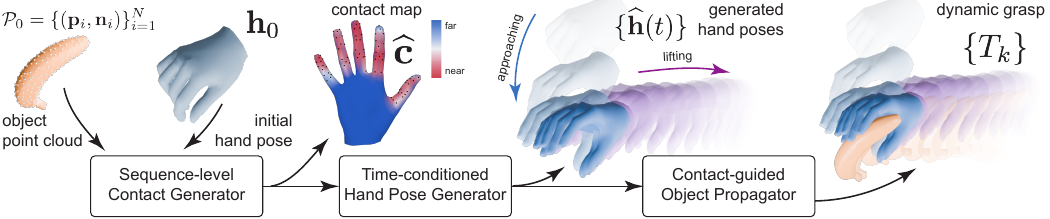}
    \caption{\textbf{Method overview.}
    Given an object point cloud $\mathcal{P}_0$ and an initial hand state $\mathbf{h}_0$, both expressed in the initial wrist frame, a contact flow samples a sequence-level contact-distance map $\widehat{\mathbf{c}}$ (Sec.~\ref{sec:contact_generation}).
    Conditioned on $\widehat{\mathbf{c}}$, a pose flow generates the hand state $\widehat{\mathbf{h}}(t)$ independently at each queried motion time $t$, sharing one noise sample across the sequence (Sec.~\ref{sec:pose_generation}).
    A geometric stage then detects the attachment frame $t_{\mathrm{att}}$ and propagates the object
rigidly with the hand, yielding transforms $\{T_k\}$, through contact-guided weighted
 registration (Sec.~\ref{sec:contactOptimization}).
    }
    \label{fig:overview}
\end{figure*}

Dynamic grasping couples two challenges.
Hand articulation is high-dimensional and under-constrained, since many grasps fit the same initial state.
By contrast, under our rigid-attachment assumption, object motion after grasp formation follows the hand.
Learning both jointly forces one network to represent a stochastic articulation prior and a deterministic rigid-body consequence.
Instead, we decompose the task into (i)~a generative model that predicts hand-side quantities and (ii)~an optimization that recovers object motion geometrically (\Cref{sec:contactOptimization}).
This confines learning to the ambiguous hand motion while allowing the geometric stage to transfer to unseen shapes.

Key to our approach, we make \textit{contact an explicit intermediate}, rather than letting the pose network implicitly discover where the hand should touch each object. Specifically, we first predict a compact, object-grounded contact target---a single sequence-level contact-distance map over the hand vertices (\Cref{sec:contact_generation})---and then generate poses conditioned on it (\Cref{sec:pose_generation}).
Both generators are trained with conditional flow matching~\cite{Lipman2023FlowMatching}, and all quantities are expressed in the canonical frame of the initial hand in the first frame.
This target lives in a hand-centric space largely invariant to object identity, turning \textit{grasp this unseen geometry} into the easier \textit{reach this contact pattern} task. We find this design choice transfers more readily across object geometry (\Cref{sec:evaluation}).

\subsection{Generating Sequence-Level Contact}
\label{sec:contact_generation}
\paragraph*{Selecting the reference contact frame.}
We identify a reference contact frame in each training sequence using the onset of recorded object motion as a proxy for contact establishment.
Let $(R_k^O,\mathbf u_k^O)$ denote the ground-truth object rotation and translation at frame $k$.
We compare consecutive frames and select the last frame before the first translation or rotation change that exceeds a threshold:
\begin{equation}
    \begin{aligned}
    \Delta p_k &=\|\mathbf u_k^O-\mathbf u_{k-1}^O\|_2,\\
    \Delta r_k &=\operatorname{angle}\!\left((R_{k-1}^O)^{\mathsf T}R_k^O\right),\\
    \text{and}\qquad g_{\mathrm{ref}} &=\min\!\left\{k:\Delta p_{k+1}>\epsilon_p\ \text{or}\
                   \Delta r_{k+1}>\epsilon_r\right\}.
    \end{aligned}
    \label{eq:training_grasp_frame}
\end{equation}
Here, $\operatorname{angle}$ returns the rotation angle in radians, and $\epsilon_p$ and $\epsilon_r$ are the translation and rotation thresholds.
The hand pose at frame $g_{\mathrm{ref}}$ defines the training contact map.
Let $\mathcal V_{g_{\mathrm{ref}}}^O=\{\mathbf s_{m,g_{\mathrm{ref}}}^O\}_{m=1}^{N_O}$ denote the object-mesh vertices at this frame. For each of the $K$ candidate MANO vertices, we measure its unsigned distance $d_j$ to the nearest object-mesh vertex and apply per-map min--max normalization:
\begin{equation}
    \begin{aligned}
    d_j&=\min_{1\leq m\leq N_O}
       \|\mathbf v_{j,g_{\mathrm{ref}}}-\mathbf s_{m,g_{\mathrm{ref}}}^O\|_2,\\
    d_{\min}&=\min_{1\leq l\leq K}d_l,
    \qquad d_{\max}=\max_{1\leq l\leq K}d_l,\\
    \text{and}\qquad
    c_j&=\begin{cases}
    (d_j-d_{\min})/(d_{\max}-d_{\min}),&d_{\max}>d_{\min},\\
    0,&d_{\max}=d_{\min}.
    \end{cases}
    \end{aligned}
    \label{eq:contact_label}
\end{equation}
The resulting contact map $\mathbf c\in[0,1]^K$ follows the fixed candidate-vertex order and encodes \textit{relative}, rather than absolute, distance within each sequence.
The $K$ candidate MANO vertices are uniformly sampled from the five finger regions.
At inference, future object motion is unavailable, so we detect attachment geometrically from the generated hand sequence (\Cref{sec:contactOptimization}).

\paragraph*{Encoding the initial configuration.}
A PointNet encoder $\Phi_c$~\cite{Qi2017PointNet} processes the $N$ XYZ samples with normals in the initial object point cloud $\mathcal P_0$.
A permutation invariant attention-pooling layer, Pooling by Multihead  Attention (PMA) \cite{Lee2019SetTransformer}, compresses the set of point features into $M_o$ object tokens, while a learned linear embedding $E_{0,c}$ maps each of the $J$ articulated-joint rotations $\mathbf q_0^{\mathrm{art}}$ to its own token:
\begin{equation}
    \mathbf g_c=\operatorname{PMA}_{o,c}(\Phi_c(\mathcal P_0))
    \quad \text{and}\quad \mathbf e_{0,c}=E_{0,c}(\mathbf q_0^{\mathrm{art}})+\mathbf B_{0,c}.
    \label{eq:object_features}
\end{equation}
These have sizes $M_o\times d$ and $J\times d$, respectively, where $d$ is the feature width.
The PMA module is a single cross-attention layer in which $M_o$ learned seed tokens act as queries over the point features, followed by a residual connection and layer normalization, and $\mathbf B_{0,c}\in\mathbb R^{J\times d}$ is a learned embedding that identifies each joint.
We omit global orientation and translation of $\mathbf h_0$ because all quantities are expressed in the canonical frame of the initial hand, where they are the identity and zero.
These per-point object features are shared by the contact and pose generators, whereas each generator learns its own pooling seeds and joint embeddings.

\paragraph*{Generating the contact map.}
Our first generator predicts contact flow  directly in the $K$-dimensional contact-map space. We sample standard Gaussian noise $\mathbf z_c\sim\mathcal N(\mathbf0,I_{K})$ and directly transport it to the ground-truth contact vector $\mathbf c$.
For contact-flow time $\tau_c\in[0,1]$, the training interpolation is
\begin{equation}
    \mathbf x_{\tau_c}^c=(1-\tau_c)\mathbf z_c+\tau_c\mathbf c.
    \label{eq:contact_path}
\end{equation}
For any scalar time $s\in[0,1]$, we concatenate the sine and cosine features at octave frequencies $2^m\pi$, for $m=0,\ldots,M_\tau-1$, into a single encoding $\gamma(s)\in\mathbb R^{2M_\tau}$~\cite{Mildenhall2020NeRF}. The learned MLP $T_{\tau,c}$ processes this complete encoding and reshapes its output into $M_\tau$ time tokens, producing $\mathbf e_{\tau_c,c}=T_{\tau,c}(\gamma(\tau_c))\in\mathbb R^{M_\tau\times d}$.
Cross-attention uses one noisy-state query token per candidate vertex. Time, object, and initial-pose tokens are concatenated with learned source embeddings:
\begin{equation}
    \begin{aligned}
    \overline{\mathbf X}_{\tau_c}^c
       &=\big[P_{\mathrm{in},c}(x_{\tau_c,j}^c)\big]_{j=1}^{K}+\mathbf B_c
         \in\mathbb R^{K\times d},\\
    \mathbf S_{\tau_c}^c
       &=\operatorname{Concat}_{\mathrm{tok}}
          (\mathbf e_{\tau_c,c},\mathbf g_c,\mathbf e_{0,c})\\
       &\qquad\in\mathbb R^{(M_\tau+M_o+J)\times d},\\
    \mathbf Q_c&=\overline{\mathbf X}_{\tau_c}^cW_{Q,c},\quad
       \mathbf K_c=\mathbf S_{\tau_c}^cW_{K,c},\\
    \text{and}\quad \mathbf V_c&=\mathbf S_{\tau_c}^cW_{V,c}.
    \end{aligned}
    \label{eq:attention_qkv}
\end{equation}
Here $P_{\mathrm{in},c}$ lifts each scalar contact value to width $d$ and $\mathbf B_c\in\mathbb R^{K\times d}$ contains the learned candidate-vertex embeddings.
The learned matrices project each feature to an attention head of width $d_a=d/H$, where $H$ is the head count.
Each of the $L_c$ blocks applies self-attention across the vertex tokens followed by this cross-attention, each with a residual connection and layer normalization, so the features remain $d$-dimensional along the residual path; a per-token feed-forward head then maps each vertex token to a scalar, producing the $K$-dimensional velocity $v_c$~\cite{Vaswani2017Attention}.
We train this field by matching the straight-line target velocity:
\begin{equation}
    \mathcal L_c=\mathbb E\left[
       \frac{1}{K}\|v_c(\mathbf x_{\tau_c}^c,\tau_c;\mathcal C_c)
                         -(\mathbf c-\mathbf z_c)\|_2^2\right],
    \label{eq:contact_loss}
\end{equation}
where $\mathcal C_c=(\mathbf g_c,\mathbf e_{0,c})$ contains the fixed conditions, and the expectation is over training sequences, noise, and flow time.

\paragraph*{Sampling the contact map.}
At inference, we integrate from Gaussian noise at $\tau_c=0$ to $\tau_c=1$:
\begin{equation}
    \frac{\mathrm d\mathbf x_{\tau_c}^c}{\mathrm d\tau_c}
       =v_c(\mathbf x_{\tau_c}^c,\tau_c;\mathcal C_c),\qquad
       \text{and}\qquad \mathbf x_0^c=\mathbf z_c.
    \label{eq:contact_sampling}
\end{equation}
We clip $\mathbf x_1^c$ element-wise to obtain the sequence-level contact map $\widehat{\mathbf c}\in[0,1]^K$.

\subsection{Generating Time-Conditioned Hand Poses}
\label{sec:pose_generation}
Our second generator predicts a hand state at any requested motion time $t\in[0,1]$, conditioned on the shared initial-state features and sequence-level contact-map values.
During training, the observed frame's normalized position within the sequence is used as $t$ and its $D_h$-dimensional MANO state $\mathbf h(t)$ as the target.
Motion time specifies the corresponding pose to generate; pose-flow time $\tau_h\in[0,1]$ specifies progress from noise to that pose.
Each pose-flow integration holds $t$ fixed while varying $\tau_h$; $\tau_c$ and $\tau_h$ are sampled independently during joint training.

\paragraph*{Reusing the conditioning features.}
The pose generator reuses the PointNet features $\Phi_c(\mathcal P_0)$ and the initial state $\mathbf h_0$, see  \Cref{sec:contact_generation}.
Following Eq.~\eqref{eq:object_features} with separate weights, it computes its own tokens $\mathbf g_h=\operatorname{PMA}_{o,h}(\Phi_c(\mathcal P_0))$ and $\mathbf e_{0,h}=E_{0,h}(\mathbf q_0^{\mathrm{art}})+\mathbf B_{0,h}$.
It is further conditioned on the contact map $\widetilde{\mathbf c}$: we use the ground-truth map $\widetilde{\mathbf c}=\mathbf c$ during training and the generated map $\widetilde{\mathbf c}=\widehat{\mathbf c}$ during inference.
A learned projection $P_c$ lifts each contact value to width $d$, and $\mathbf B_{c,h}\in\mathbb R^{K\times d}$ identifies its candidate vertex.
We use a second PMA layer to pool the resulting $K$ tokens into $M_c$ contact tokens:
\begin{equation}
    \overline{\mathbf C}=\big[P_c(\widetilde c_j)\big]_{j=1}^{K}+\mathbf B_{c,h},
    \qquad \mathbf e_c=\operatorname{PMA}_{c,h}(\overline{\mathbf C}).
    \label{eq:pose_contact_tokens}
\end{equation}
This projection embeds contact conditions, not a generative latent space.
Ground-truth contact maps come directly from the data loader, so pose-loss gradients update the shared PointNet encoder but not the contact flow.
PointNet features are computed once for both generators.
Separate MLPs project the octave sine--cosine encodings of pose-flow time and motion time into $\mathbf e_{\tau_h,h}=T_{\tau,h}(\gamma(\tau_h))$ and $\mathbf e_t=T_t(\gamma(t))$, respectively, each comprising $M_\tau$ tokens.

\paragraph*{Generating the time-conditioned pose.}
We sample standard Gaussian noise $\mathbf z_h\sim\mathcal N(\mathbf0,I_{D_h})$ and form the training interpolation
\begin{equation}
    \begin{aligned}
    \boldsymbol\delta_h(t)&=\mathbf h(t)-\mathbf h_0,\\
    \mathbf y_h(t)&=(\boldsymbol\delta_h(t)-\boldsymbol\mu)\oslash\boldsymbol\sigma,\\
    \text{and}\quad
    \mathbf x_{\tau_h}^h(t)&=(1-\tau_h)\mathbf z_h+\tau_h\mathbf y_h(t).
    \end{aligned}
    \label{eq:pose_path}
\end{equation}
Here, $\oslash$ is element-wise division. We compute one scalar mean and standard deviation for each of the global-orientation, articulation, and translation blocks of the raw offsets $\boldsymbol\delta_h(t)$ over the training set, and repeat these block-wise statistics to form $\boldsymbol\mu,\boldsymbol\sigma\in\mathbb R^{D_h}$. Thus, $\mathbf y_h(t)$ is the normalized pose offset predicted by the flow.
A learned linear map $E_h$ splits this noisy state into $J+1$ rotation-parameter blocks and one translation block and maps each to its own query token, with a learned embedding identifying each token, producing $\overline{\mathbf X}_{\tau_h}^h(t)=E_h(\mathbf x_{\tau_h}^h(t))\in\mathbb R^{(J+2)\times d}$.
As in the contact generator, the conditioning features are concatenated along the token dimension, each offset by a learned embedding of its source, now including the projected contact map and motion time:
\begin{equation}
    \begin{aligned}
    \mathbf S_{\tau_h,t}^h&=\operatorname{Concat}_{\mathrm{tok}}
       (\mathbf g_h,\mathbf e_{0,h},\mathbf e_c,\mathbf e_{\tau_h,h},\mathbf e_t)\\
       &\qquad\in\mathbb R^{(M_o+J+M_c+2M_\tau)\times d},\\
    \mathbf Q_h&=\overline{\mathbf X}_{\tau_h}^h(t)W_{Q,h},\quad
    \mathbf K_h=\mathbf S_{\tau_h,t}^hW_{K,h},\\
    \text{and}\quad \mathbf V_h&=\mathbf S_{\tau_h,t}^hW_{V,h}.
    \end{aligned}
    \label{eq:pose_attention_qkv}
\end{equation}
The attention stack follows the contact generator, with separate weights and $L_h$ blocks, followed by a residual feed-forward layer and per-token heads. Each rotation token outputs $d_R$ velocity components and the translation token outputs three, forming the $D_h$-dimensional velocity $v_h$ in normalized pose-offset space.
The $d$-dimensional query embeddings are internal features; the flow state remains in the normalized MANO parameter space.
We train the velocity field with the loss term,
\begin{equation}
    \mathcal L_h=\mathbb E\Big[\sum_{b}\tfrac{1}{|b|}\big\|\big[
      v_h(\mathbf x_{\tau_h}^h(t),\tau_h;\mathcal C_h,t)
      -(\mathbf y_h(t)-\mathbf z_h)\big]_b\big\|_2^2\Big],
    \label{eq:pose_loss}
\end{equation}
where $b$ ranges over the global-orientation, articulation, and translation blocks of sizes $d_R$, $Jd_R$, and 3, so that each block contributes equally regardless of its dimension.
Here, $\mathcal C_h=(\mathbf g_h,\mathbf e_{0,h},\mathbf e_c)$ collects the object, initial-hand, and contact-map features.
The expectation is over training frames, Gaussian noise, and flow time.
Teacher forcing avoids sampling the contact flow for this loss.

\paragraph*{Recovering the hand sequence.}
At inference, we construct $\widehat{\mathbf e}_c$ by replacing $\widetilde{\mathbf c}$ with $\widehat{\mathbf c}$ in Eq.~\eqref{eq:pose_contact_tokens} and set $\widehat{\mathcal C}_h=(\mathbf g_h,\mathbf e_{0,h},\widehat{\mathbf e}_c)$. We then integrate the pose flow for each requested motion time and map its terminal state back to MANO parameters:
\begin{equation}
    \begin{aligned}
    \frac{\mathrm d\mathbf x_{\tau_h}^h(t)}{\mathrm d\tau_h}
        &=v_h(\mathbf x_{\tau_h}^h(t),\tau_h;
                 \widehat{\mathcal C}_h,t),\\
    \mathbf x_0^h(t)&=\mathbf z_h,
    \qquad \text{and}\\
    \widehat{\mathbf h}(t)&=\mathbf h_0+\boldsymbol\mu
        +\boldsymbol\sigma\odot\mathbf x_1^h(t).
    \end{aligned}
    \label{eq:pose_sampling}
\end{equation}
We evaluate MANO with $\widehat{\mathbf h}(t)$ and the fixed shape coefficients to recover the hand mesh at each time.
The same generated contact map $\widehat{\mathbf c}$ and pose-noise sample $\mathbf z_h$ are reused across all motion-time queries in one sequence.
This couples the stochastic inputs without autoregressive rollout.

\begin{table*}[!t]
\centering
\caption{
\textbf{Quantitative comparison.}
Sequence-mean results on DexYCB unseen objects (200 sequences) and Open-CHOIR (30). LatentHOI and ours select one of ten samples using the attachment and equal-weight HO-Dist/IV ranking protocol. Single-sample timings use the same GPU: LatentHOI generates 96 frames with 1{,}000 DDPM steps; ours generates 64 with 32 Euler steps.}
\small
\setlength{\tabcolsep}{3.5pt}
\begin{tabular}{llccccccccc}
\toprule
 & & \multicolumn{2}{c}{Contact} & \multicolumn{3}{c}{Penetration} & \multicolumn{2}{c}{Smoothness} & \multicolumn{2}{c}{Generation time} \\
\cmidrule(lr){3-4}\cmidrule(lr){5-7}\cmidrule(lr){8-9}\cmidrule(lr){10-11}
Dataset & Method & HO-Dist$\downarrow$ & HO-Dist$_{778}\downarrow$ & IV$\downarrow$ & PR$\downarrow$ & ID$\downarrow$ & Acc$_h\downarrow$ & Acc$_o\downarrow$ & Seq.$\downarrow$ & Frame$\downarrow$ \\
 & & (cm) & (cm) & (cm$^3$) & (\%) & (mm) & (cm/frame$^2$) & (cm/frame$^2$) & (s) & (ms) \\
\midrule
\multirow{3}{*}{DexYCB}
 & LatentHOI~\cite{Li2025LatentHOI} & \textbf{0.090} & \textbf{0.056} & 1.72 & 3.80 & 10.3 & 0.587 & 0.330 & 8.24 & 85.8 \\
 & Ours (GT init.) & 0.105 & 0.065 & \textbf{1.59} & \textbf{3.30} & \textbf{8.7} & 0.451 & \textbf{0.291} & \textbf{0.18} & \textbf{2.8} \\
 & Ours (LatentHOI init.) & 0.092 & 0.059 & 2.40 & 4.66 & 10.5 & \textbf{0.363} & 0.297 & \textbf{0.18} & \textbf{2.8} \\
\midrule
\multirow{3}{*}{Open-CHOIR}
 & LatentHOI~\cite{Li2025LatentHOI} & 0.145 & 0.085 & 1.70 & 3.22 & 8.3 & 0.615 & 0.364 & 8.24 & 85.8 \\
 & Ours (GT init.) & 0.159 & 0.089 & 1.81 & 3.38 & 7.8 & \textbf{0.370} & \textbf{0.302} & \textbf{0.18} & \textbf{2.8} \\
 & Ours (LatentHOI init.) & \textbf{0.139} & \textbf{0.077} & \textbf{1.69} & \textbf{3.09} & \textbf{6.1} & 0.379 & 0.321 & \textbf{0.18} & \textbf{2.8} \\
\bottomrule
\end{tabular}
\label{tab:comparison}
\end{table*}

\subsection{Contact-Guided Object Propagation}
\label{sec:contactOptimization}

\paragraph*{Detecting the attachment frame.}
We compare each generated hand mesh with the object at its initial pose.
For frame $k$, let $\rho_k^{\mathrm{pen}}$ denote the penetration ratio, let $n_k(\epsilon_c)$ denote the number of hand vertices whose nearest-object-vertex distance is less than $\epsilon_c$, and let $D_k^{\mathrm{HO}}(k_D)$ denote the mean distance of the $k_D$ closest hand--object vertex pairs.
We measure penetration ratio over all the MANO vertices, $n_k(\epsilon_c)$ over a dense set of $K_f$ vertices covering the five finger regions, and $D_k^{\mathrm{HO}}(k_D)$ over the $K$ contact candidates (see \Cref{sec:contact_generation}), which sparsely sample the same regions.
We first define the set of geometrically valid attachment-frame candidates:
\begin{equation}
    \mathcal I_{\mathrm{valid}}
    =\left\{k:\rho_k^{\mathrm{pen}}\leq\epsilon_{\mathrm{pen}},\;
                 n_k(\epsilon_c)\geq n_{\min}\right\}.
    \label{eq:valid_grasp_frames}
\end{equation}
Among these candidates, we select the frame with the closest hand--object approach:
\begin{equation}
    g_{\mathrm{att}}
    =\mathop{\mathrm{arg\,min}}_{k\in\mathcal I_{\mathrm{valid}}}
       D_k^{\mathrm{HO}}(k_D)
    \qquad \text{and}\qquad t_{\mathrm{att}}=t_{g_{\mathrm{att}}}.
    \label{eq:grasp_time}
\end{equation}
If $\mathcal I_{\mathrm{valid}}=\varnothing$, no attachment is detected and the object remains at its initial pose throughout the sequence.
\paragraph*{Propagating the object.}
At $g_{\mathrm{att}}$, we select $K_a$ attachment points and assign nonnegative confidence weights to their hand--object correspondences.
The attachment points are chosen by furthest-point sampling over the vertices of the $K_f$-vertex set that lie within $\epsilon_c$ of the object at $g_{\mathrm{att}}$, seeded at the vertex closest to the object, so that they spread over the contact region.
Each attachment vertex $\mathbf v_{i,g_{\mathrm{att}}}$ is paired with a distinct nearest object-mesh vertex $\mathbf s_i^O$, and we store the offset $\mathbf o_i=\mathbf v_{i,g_{\mathrm{att}}}-\mathbf s_i^O$ and the weight $w_i=\exp(-(d_i/\sigma_w)^2)$, where $d_i$ is the distance from $\mathbf v_{i,g_{\mathrm{att}}}$ to the object and $\sigma_w=\epsilon_c/2$.
For each subsequent frame, a contact-guided weighted registration estimates an incremental rigid transform $\Delta T_k\in\mathrm{SE}(3)$ from these attachment points, and the cumulative object transform is updated as $T_k=\Delta T_kT_{k-1}$:
\begin{equation}
    \Delta T_k=\mathop{\mathrm{arg\,min}}_{\Delta T\in\mathrm{SE}(3)}
       \sum_{i=1}^{K_a}w_i\,\eta_{i,k}\,
       \big\|\Delta T\,T_{k-1}\mathbf s_i^O
       -(\mathbf v_{i,k}-R_{k-1}\mathbf o_i)\big\|_2^2.
    \label{eq:object_registration}
\end{equation}
Here, $\mathbf v_{i,k}$ is the attachment vertex at frame $k$ and $R_{k-1}$ is the rotation of $T_{k-1}$, so subtracting $R_{k-1}\mathbf o_i$ preserves the hand--object gap observed at $g_{\mathrm{att}}$.
Each fit is solved in closed form by weighted Kabsch alignment and refined over $L_{\mathrm{IRLS}}$ iteratively reweighted passes, where $\eta_{i,k}=\min(1,\delta/r_{i,k})$ is a Huber weight on the current residual $r_{i,k}$ that softly down-weights a finger that slips or lifts.
Note that the correspondences are fixed at $g_{\mathrm{att}}$; only the weights $\eta_{i,k}$ are re-estimated.
We set $T_k=I$ for the frames $k\leq g_{\mathrm{att}}$, preserving the initial object placement until the attachment frame.
This training-free stage keeps contacts fixed and assumes sustained attachment; release and sliding are not modeled.

\section{Evaluation}
\label{sec:evaluation}

\begin{figure*}[!tp]
    \centering
    \includegraphics[width=0.97\textwidth]{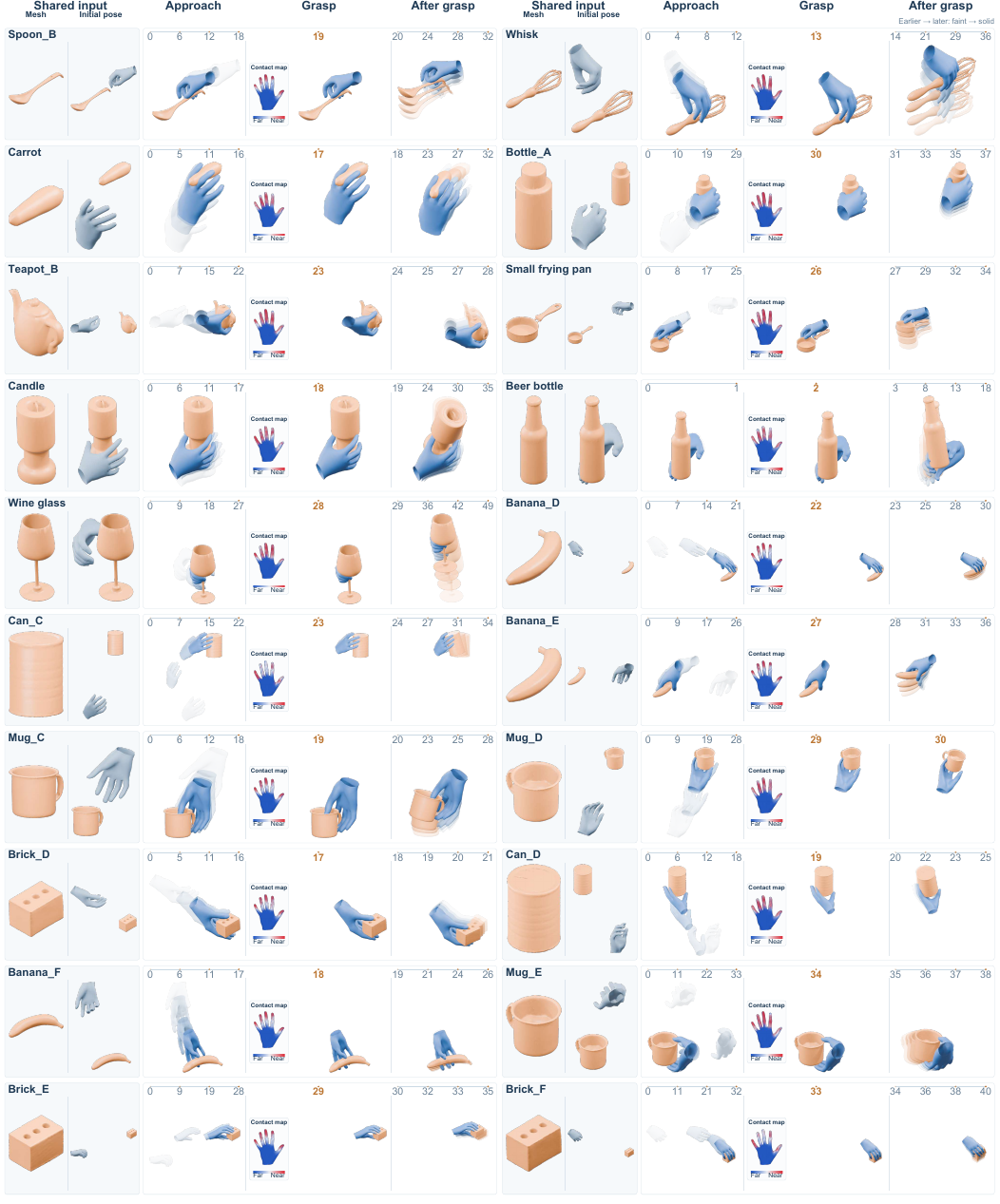}
    \caption{\textbf{Results gallery.}
    Generated motions for 20 scenes: input mesh, initial pose, approach, grasp, and post-grasp stages. Increasing opacity indicates temporal order; numbers identify frames. Insets show predicted contact maps on the rest-pose hand.}
    \label{fig:gallery}
\end{figure*}

\begin{figure*}[!tp]
    \centering
    \includegraphics[width=0.97\textwidth]{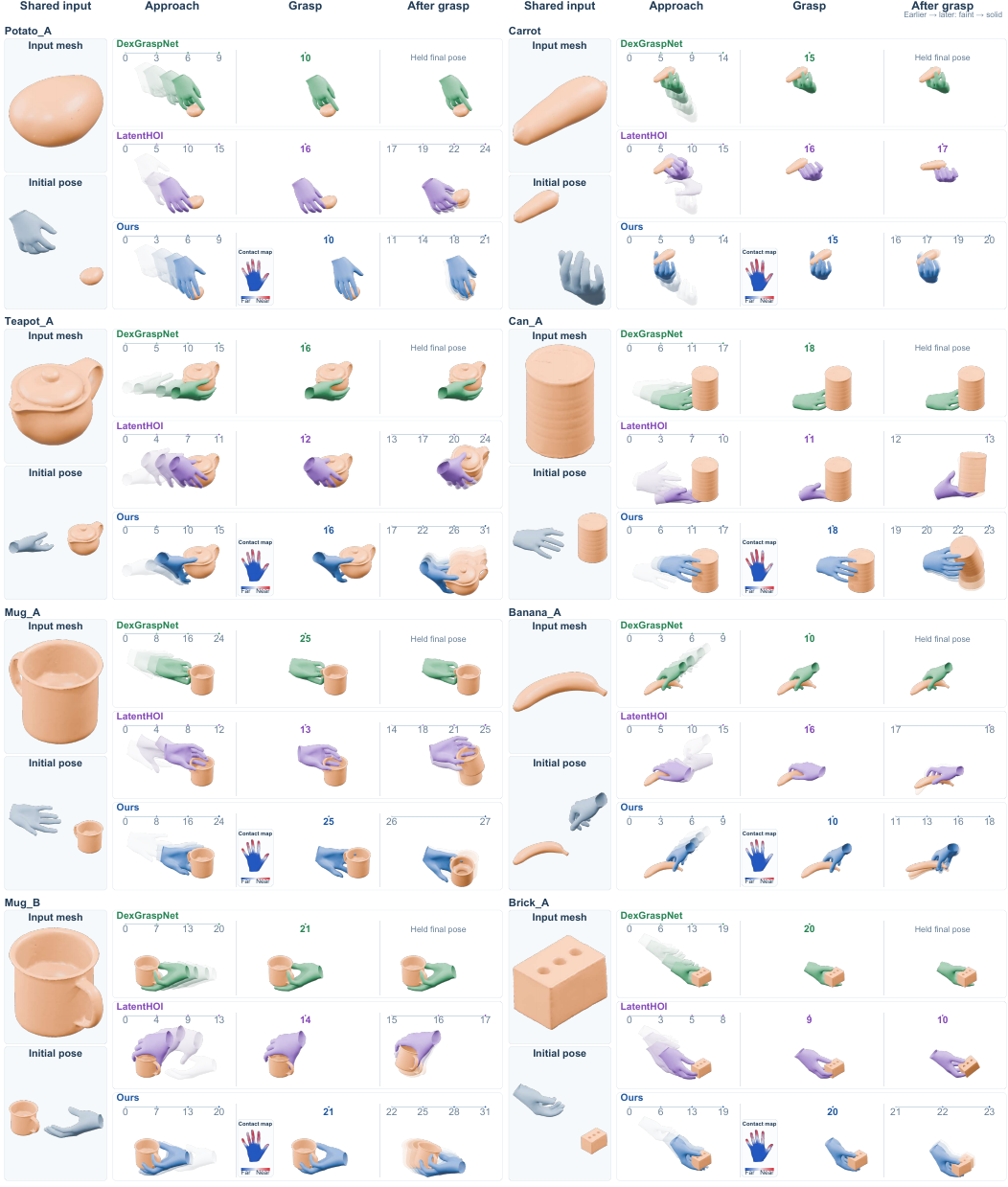}
    \caption{\textbf{Comparison.} Evaluation on eight DexYCB and Open-CHOIR scenes. DexGraspNet interpolates to a static grasp and holds the final poses; LatentHOI and ours generate motion. Opacity indicates temporal order; numbers identify frames. Insets show our contact maps.}
    \label{fig:comparison}
\end{figure*}

\begin{figure*}[!t]
    \centering
    \includegraphics[width=\textwidth]{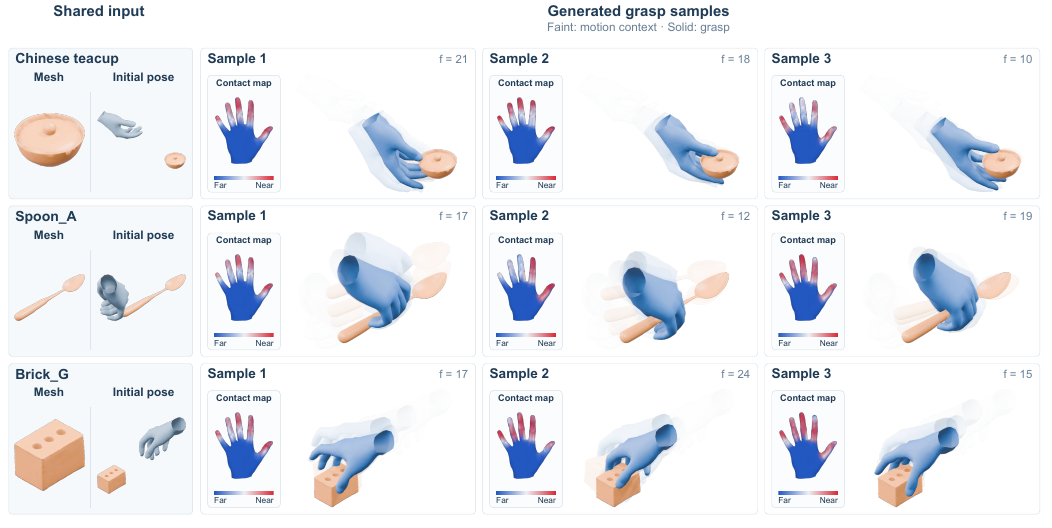}
    \caption{ \textbf{Sampling multiple variations.}
    Three generated samples for the same relative initial configuration for different objects. Solid poses show the grasp frame; translucent poses show preceding and subsequent motion. Insets show predicted contact maps.}
    \label{fig:diversity}
\end{figure*}

\paragraph*{Datasets.}
DexYCB~\cite{Chao2021DexYCB} contains 1,000 grasping sequences across 20 YCB object categories. Following LatentHOI~\cite{Li2025LatentHOI}, we train on 799 sequences of 16 objects and evaluate on 200 sequences of four held-out objects: \texttt{master chef can}, \texttt{banana}, \texttt{mug}, and \texttt{foam brick}. One sequence lacks complete hand annotations and is excluded. Left-hand sequences are mirrored to the right hand. For mesh reconstruction, we use the annotated MANO shape coefficients. We call this protocol \emph{unseen object}.
For \emph{in-the-wild unseen object} evaluation, we use Open-CHOIR~\cite{Xu2026OpenCHOIR} reconstructions with novel object geometries, without fine-tuning.
The 30 test sequences each show a right hand manipulating a reconstructed object.
The Open-CHOIR hand and object parameters are converted to the DexYCB annotation format by folding the release's hand-only frame rotation into the MANO global orientation and baking the per-axis object scale into the rest mesh, and each sequence is then expressed in the canonical frame of its initial hand, as for DexYCB.
Since the release provides no shape coefficients, the hand uses the MANO mean shape.
Reconstructions serve as pseudo-ground truth. We retain the interval with both hand and object parameters and evaluate generated motions at 65 uniformly spaced normalized sequence times.
\paragraph*{Comparison.}
We train LatentHOI~\cite{Li2025LatentHOI} on the same split using its official single-hand implementation and pretrained GraspVAE. It is a task-level, not input-matched, baseline: it receives canonical object geometry and the prompt ``pick up $\langle$object$\rangle$'' (``pick up object'' for Open-CHOIR), but no initial hand pose or reference trajectory. Its default 1{,}000-step DDPM sampler generates 96 frames.

\emph{Ours (GT init.)} uses the ground-truth first-frame hand pose; \emph{Ours (LatentHOI init.)} uses LatentHOI's generated first frame. Both methods use the same geometric object propagation and 65-frame evaluation grid, resampling LatentHOI outputs at the corresponding normalized times. For each sequence and method, we draw 10 samples and rank those with finite HO-Dist and IV by $\operatorname{rank}(\text{HO-Dist})+\operatorname{rank}(\text{IV})$. We select the highest-ranked sample with a detected attachment, or the highest-ranked valid sample if none has an attachment. This preference tests geometric attachment, not physical lifting (details in \Cref{sec:supp_baseline}). Both methods use all 200 DexYCB and 30 Open-CHOIR test sequences.

The DexGraspNet~\cite{Wang2023DexGraspNet} baseline uses its MANO implementation to synthesize a static grasp and interpolates from \emph{LatentHOI's initial hand pose} during approach. This tests whether endpoint interpolation suffices for realistic grasping motion. Since it does not generate post-grasp motion, we include it in the qualitative comparison and user study, but not the full-sequence metrics in Table~\ref{tab:comparison}.

\paragraph*{Metrics.}
Following LatentHOI~\cite{Li2025LatentHOI} and Open-CHOIR~\cite{Xu2026OpenCHOIR}, HO-Dist and HO-Dist$_{778}$ measure closest hand--object approach using contact candidates and all 778 MANO vertices, respectively. Interpenetration Volume (IV, cm$^3$) measures hand volume inside the object; Penetration Ratio (PR, \%) is the fraction of penetrating hand vertices; Interpenetration Depth (ID, mm) is the maximum depth. Acc$_h$ and Acc$_o$ measure hand and object smoothness using second-finite-difference magnitudes (cm/frame$^2$). Lower is better for all metrics. We aggregate over each 65-frame sequence using recovered object motion, then report means across sequences. \Cref{sec:supp_metrics} gives calculation details.

\paragraph*{Implementation details.}
\label{sec:implementation}
\textit{Architecture.}
We use $N=1{,}024$ object points, $K=75$ contact candidates, $K_f=245$ dense finger-region vertices, and $K_a=8$ attachment points. The feature width is $d=256$, with $M_o=M_c=16$ pooled object and contact tokens, $M_\tau=8$ time tokens, $L_c=4$ contact-flow blocks, $L_h=6$ pose-flow blocks, and $L_{\mathrm{IRLS}}=3$ robust-registration passes. Both generators use $H=8$ attention heads, feed-forward width 1{,}024, and dropout 0.1. The PointNet encoder $\Phi_c$ uses pointwise widths 64, 128, and 1{,}024, followed by a linear projection to $d$, and each time MLP has one hidden layer of width $d$.

\textit{Training.}
Both flows are trained concurrently for 300 epochs using AdamW with a constant learning rate and weight decay of $10^{-4}$ and batch size 64, using every observed frame as a target and sampling the two flow times independently from $\mathcal U(0,1)$. We keep the checkpoint with the lowest sampled generation error on a
validation subset of $20$ training sequences. Each epoch, every training sequence is re-anchored at a sampled pre-contact frame, with probability 0.4 of retaining the original first frame and a quadratic bias toward earlier frames otherwise. The input object cloud is redrawn from a four-times-larger farthest-point-sampled pool and perturbed with 1\,mm Gaussian jitter. A single AdamW optimizer minimizes $\mathcal L_h$ at every mini-batch and adds $\mathcal L_c$ once every 32 mini-batches. The active terms have equal coefficients, and the three blocks of \Cref{eq:pose_loss} are equally weighted. The pose branch and shared PointNet receive gradients at every step; the contact branch receives gradients only when $\mathcal L_c$ is included.

\textit{Inference and post-processing.}
We use $\epsilon_p=1$\,cm and $\epsilon_r=\pi/36$ rad ($5^\circ$) to select the training contact frame, and $\epsilon_{\mathrm{pen}}=0.08$, $\epsilon_c=5$\,mm, $n_{\min}=8$, $k_D=10$, and $\delta=1$\,cm for post-processing. Both flows use 32 forward-Euler steps. We evaluate sequences of $T=65$ frames, retaining the initial frame at $t_0=0$ and generating 64 frames at $t_k=k/64$ for $k=1,\ldots,64$. One contact map is sampled per sequence, and the same pose-noise sample is shared across all motion-time queries.

\paragraph*{Runtime.}
The contact flow is integrated once per sequence and the pose flow at all 64 requested times in parallel.
Attachment detection and object propagation follow hand generation.
We time 200 DexYCB sequences on an NVIDIA RTX 4070 Ti SUPER, each with 65 frames and 32 Euler steps per flow, synchronizing the GPU around generation and excluding the first sequence as warm-up.
Contact and hand generation take 0.18\,s per 65-frame sequence, or 2.8\,ms per generated frame. We report this fixed-length generation time for both DexYCB and Open-CHOIR in Table~\ref{tab:comparison}. Total time including data loading is 0.50\,s per sequence.

\paragraph*{Qualitative evaluation.}
We examine motion quality, comparison with baselines, and sampling diversity. Figure~\ref{fig:gallery} shows approach, finger closure (attachment), and post-attachment contact on unseen DexYCB objects under both initializations and on in-the-wild Open-CHOIR objects. Figure~\ref{fig:comparison} compares these motion stages with LatentHOI and DexGraspNet on eight scenes; this is a task-level comparison, since LatentHOI does not use the observed initial configuration; \Cref{sec:supp_qual} shows eight further scenes. Finally, Figure~\ref{fig:diversity} shows three samples for each of three shared initial configurations, illustrating variation in generated dynamic grasps.

\paragraph*{User study.}
Fifteen participants evaluated the same sixteen scenes: ten DexYCB and six Open-CHOIR. Ours, LatentHOI, and DexGraspNet were shown anonymously as A/B/C, with randomized columns and scene order. Two synchronized views (overall and side/contact) exposed contact and penetration; viewpoints, frame rate, rendering, and playback loops were identical across methods.

\begin{table}[h!]
    \centering
    \caption{
    \textbf{User study.}
    Preferences of 15 participants over 16 scenes. Higher percentages indicate greater preference.}
    \label{tab:user_study}
    \small
    \setlength{\tabcolsep}{3pt}
    \begin{tabular}{lcc}
    \toprule
    Method & \shortstack{Naturalness /\\temporal coherence} & \shortstack{Grasp plausibility /\\apparent stability} \\
    \midrule
    DexGraspNet & 11.67\% & 10.42\% \\
    LatentHOI & 22.08\% & 25.00\% \\
    Ours & \textbf{66.25\%} & \textbf{64.58\%} \\
    \bottomrule
    \end{tabular}
\end{table}

Participants answered two three-way forced-choice questions: ``Which motion appears the most natural and temporally coherent?'' and ``Which motion results in the most plausible and stable-looking grasp?'' Ours received 66.25\% and 64.58\% preference, respectively (Table~\ref{tab:user_study}). These judgments assess perceived rather than physically verified grasp quality; we next report complementary geometric metrics.

\begin{table}[b!]
    \centering
    \caption{
    \textbf{Ablations.}
    Contact-conditioning ablation on DexYCB with GT initialization for all variants (sequence means; lower is better): sampled map (full model), ground-truth map (oracle), and no contact branch or conditioning (w/o map).}
    \label{table:ablation}
    \small
    \setlength{\tabcolsep}{3pt}
    \begin{tabular}{lccc}
    \toprule
    Metric$\downarrow$ & Full model & GT map & w/o map \\
    \midrule
    HO-Dist (cm) & 0.105 & \textbf{0.102} & 0.108 \\
    HO-Dist$_{778}$ (cm) & 0.065 & \textbf{0.062} & 0.071 \\
    IV (cm$^3$) & \textbf{1.59} & 1.85 & 1.79 \\
    PR (\%) & \textbf{3.30} & 3.66 & 3.60 \\
    ID (mm) & \textbf{8.7} & 10.0 & 10.4 \\
    Acc$_h$ (cm/frame$^2$) & 0.451 & 0.450 & \textbf{0.397} \\
    Acc$_o$ (cm/frame$^2$) & 0.291 & 0.284 & \textbf{0.237} \\
    \bottomrule
    \end{tabular}
\end{table}

\paragraph*{Quantitative evaluation.}
Table~\ref{tab:comparison} reports contact, penetration, and smoothness. Contact and penetration must be interpreted together: a hand that never reaches the object can have low penetration.
On DexYCB, LatentHOI achieves the closest contact, while ours with GT initialization has the lowest penetration and smoothest object motion. LatentHOI initialization brings our contact distances closer to LatentHOI and yields the smoothest hand motion, but increases penetration. On Open-CHOIR, ours with LatentHOI initialization has the best contact and penetration scores; GT initialization gives the smoothest hand and object motion. Our sampler is substantially faster on both datasets. These results demonstrate transfer to unseen reconstructed objects, with sensitivity to initialization.

\paragraph*{Ablations.}
Table~\ref{table:ablation} shows that predicting a contact map reduces PR from 3.60\% without contact conditioning to 3.30\% with the full model. The ground-truth map improves contact proximity but not penetration, indicating that proximity alone does not ensure a better grasp.

\begin{figure}[h!]
    \centering
    \includegraphics[width=\columnwidth,height=0.40\textheight,keepaspectratio]{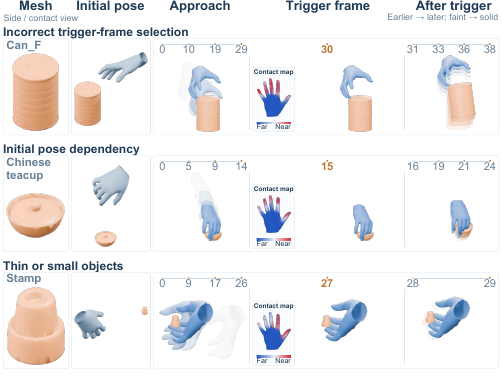}
    \caption{\textbf{Limitations.} Failures from incorrect trigger-frame selection (top), initial-pose sensitivity (middle), and thin or small objects (bottom). Rows show input, approach, trigger frame, and subsequent motion; insets show contact maps. See text for details. }
    \label{fig:failure_cases}
    \vspace{-2mm}
\end{figure}

\paragraph*{Failure cases and limitations.}
Figure~\ref{fig:failure_cases} illustrates trigger-frame errors, initial-pose sensitivity, and failures on thin or small objects. Thin, concave, or noisy geometry can cause incomplete closure or penetration; incorrect attachment frames or correspondences can produce object drift or implausible coupling. Fixed rigid attachment excludes sliding, regrasping, release, and articulated or deformable objects. The geometric criterion does not guarantee stability under gravity or external forces. Physics-aware refinement and adaptive correspondences are promising extensions.

\section{Conclusion}
\label{sec:conclusion}

We presented a hand-centric framework that generates complete grasping sequences from a single initial hand--object configuration. Conditional flow matching predicts sequence-level contact and time-indexed hand poses; contact-guided registration then recovers rigid-object motion without further learning. Evaluation on unseen DexYCB objects and in-the-wild Open-CHOIR reconstructions demonstrates transferable hand dynamics, while contact quality and attachment remain challenging under geometric distribution shift. We reported better quality as well as significant speedup over existing alternatives. Physics-aware contact refinement, adaptive correspondence updates, and richer object dynamics could extend this approach beyond rigid pick-up motions.

{\small
\bibliographystyle{abbrvnat}
\bibliography{\papersrc/dynamicgrasping}
}

\twocolumn[{%
\begin{center}
  {\Large\bfseries Supplementary Material}
\end{center}
\vspace{2ex}
}]
\appendix
\section{Evaluation metrics}
\label{sec:supp_metrics}
The definitions below expand the metric descriptions in \Cref{sec:evaluation}. All methods use the same metric computations on $T=65$ uniformly spaced normalized sequence times, comprising the initial frame and 64 subsequent frames. Each metric is aggregated within a sequence and then averaged across sequences. Contact and penetration are measured against the recovered object trajectory.

\paragraph*{Contact proximity.}
Let $\mathbf v_{j,k}$ be hand vertex $j$ at frame $k$, and let $\mathbf s_{m,k}^O$ be object-mesh vertex $m$. For a hand-vertex set $\mathcal H$ and $N_O$ object vertices, form all pairwise distances
\begin{equation}
 d_{jm,k}=\|\mathbf v_{j,k}-\mathbf s_{m,k}^O\|_2,
 \qquad j\in\mathcal H,\quad m=1,\ldots,N_O.
\end{equation}
Let $d_{(i),k}$ denote the $i$th smallest entry across all $|\mathcal H|N_O$ pairs. The sequence score is
\begin{equation}
 D^{\mathrm{HO}}(\mathcal H)
 =\min_{0\leq k<T}\frac{1}{k_D}\sum_{i=1}^{k_D}d_{(i),k},
 \qquad k_D=10.
\end{equation}
HO-Dist uses the $K=75$ contact candidates, whereas HO-Dist$_{778}$ uses all 778 vertices of the MANO hand model~\cite{Romero2017MANO}. Both are reported in centimetres. The selected pairs may share a hand vertex; this differs from averaging one nearest-object distance for each of ten distinct hand vertices. Taking the minimum over time measures closest approach rather than sustained contact, so these scores are interpreted together with penetration and qualitative motion results.

\paragraph*{Interpenetration volume.}
IV is the mean hand--object overlap volume over frames, estimated with 0.5\,cm voxels and reported in cm$^3$. We use Trimesh for inside/outside and surface-distance queries, treating the object meshes as watertight.

\paragraph*{Penetration ratio and depth.}
Let $s_{j,k}$ be the object's signed distance at MANO vertex $j$, positive inside, and let $b_{j,k}=\mathbf 1[s_{j,k}>0]$. Penetration ratio uses the original 778 vertices:
\begin{equation}
 \mathrm{PR}=\frac{100}{T}\sum_{k=0}^{T-1}
 \frac{1}{778}\sum_{j=1}^{778}b_{j,k}.
\end{equation}
The interpenetration depth is the deepest penetrating vertex over the sequence:
\begin{equation}
 \mathrm{ID}=\max_{0\leq k<T}\max_{1\leq j\leq778}\max(0,s_{j,k}).
\end{equation}
ID measures distance to the triangle surface and is zero without penetration. It is reported in millimetres and takes the maximum over frames, unlike the frame-averaged IV and PR.

\paragraph*{Temporal smoothness.}
Let $\mathbf a_{j,k}$ be the position of hand joint $j$, and let $\mathbf o_k$ be the arithmetic mean of the posed object-mesh vertices, both in centimetres. We use second differences in frame index:
\begin{align}
 \mathrm{Acc}_h&=\frac{1}{21(T-2)}\sum_{k=1}^{T-2}\sum_{j=1}^{21}
 \|\mathbf a_{j,k+1}-2\mathbf a_{j,k}+\mathbf a_{j,k-1}\|_2,\\
 \mathrm{Acc}_o&=\frac{1}{T-2}\sum_{k=1}^{T-2}
 \|\mathbf o_{k+1}-2\mathbf o_k+\mathbf o_{k-1}\|_2.
\end{align}
These scores are expressed in cm/frame$^2$, without conversion to physical acceleration. Smoothness should be interpreted together with contact and object motion.

\section{Baseline adaptation and sample selection}
\label{sec:supp_baseline}
LatentHOI~\cite{Li2025LatentHOI} produces 96-frame motions conditioned on canonical object geometry and a pick-up prompt. Its output is resampled to the common 65-frame evaluation grid. Our method is evaluated from either the observed initial hand pose or a LatentHOI-generated initial pose, as described in \Cref{sec:evaluation}.

For each sequence, LatentHOI and our method each draw ten samples and use the same selection procedure. Samples with non-finite HO-Dist or IV are excluded. Among the remaining samples, HO-Dist and IV are ranked separately in ascending order, starting at one; exact ties receive their mean rank. Ranks are computed over this entire valid population, before applying the attachment preference. Samples are ordered by the sum of the two ranks, then by HO-Dist, then IV, with remaining ties retaining generation order.

We select the first sample in that order with at least one attached frame. The attachment flag is set at the geometrically detected trigger frame and remains set thereafter. This tests geometric attachment rather than physical lifting. If no sample is attached, the top-ranked valid sample is retained; if no valid sample exists, the sequence is excluded from the selected-sample average. Dataset scores average the selected sample's metrics equally across sequences.

\begin{figure*}[!p]
    \centering
    \includegraphics[width=0.97\textwidth]{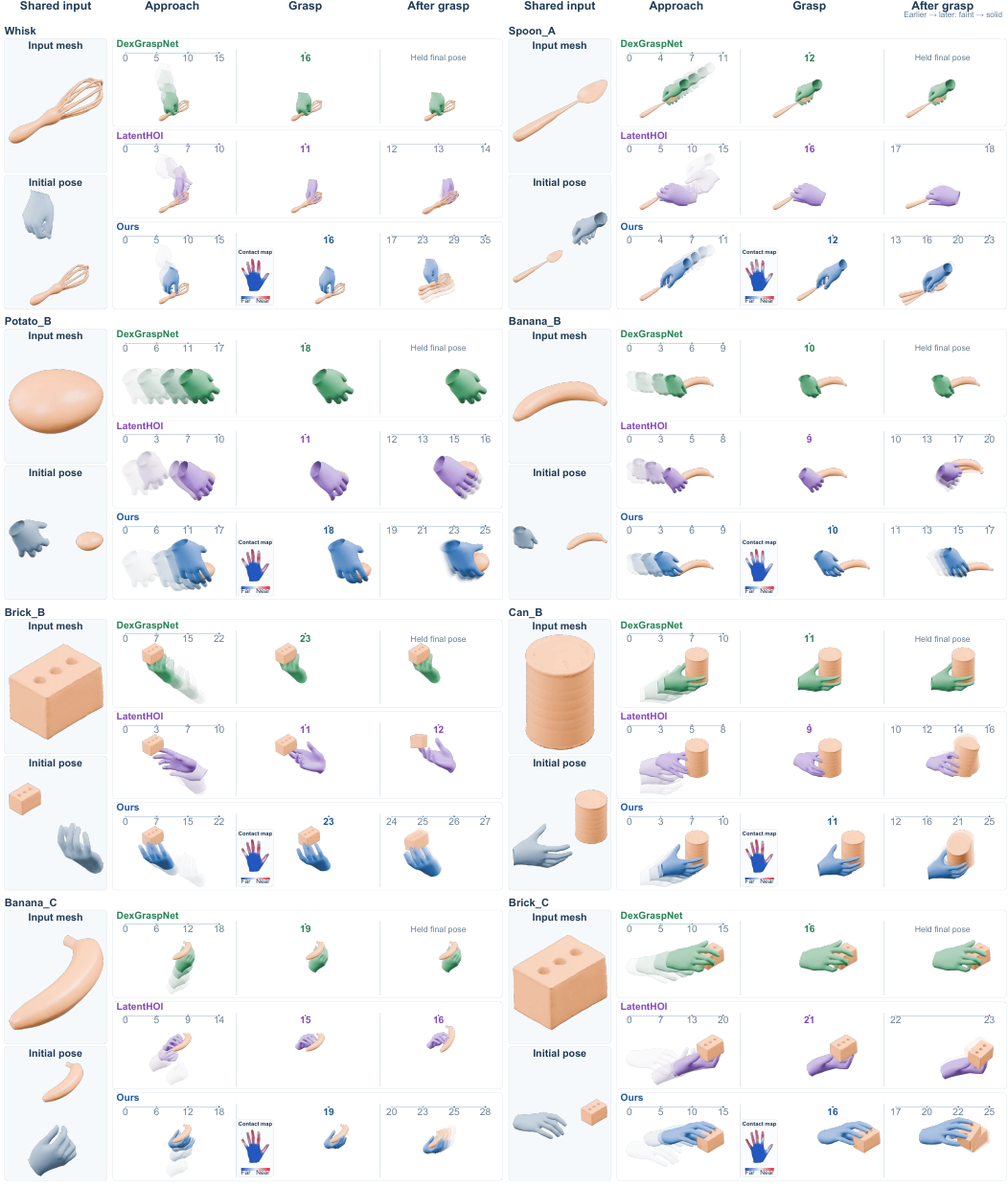}
    \caption{Additional qualitative comparisons on eight scenes. Translucent hands indicate earlier frames, opaque hands later frames, and numbers identify frames. DexGraspNet interpolates to a static grasp and holds its final pose for display; LatentHOI and ours generate subsequent motion. Insets show our predicted contact maps.}
    \label{fig:supp_comparison}
\end{figure*}

\section{Additional implementation details}
\label{sec:supp_impl}
Pose offsets use the continuous 6D rotation representation~\cite{Zhou2019Rotation} and are normalized separately for global orientation, articulation, and translation. Each block has one scalar mean and standard deviation, repeated across the parameters in that block. The initial articulation uses a separate scalar mean and standard deviation. Object XYZ coordinates use a per-axis mean and one shared standard deviation; normals are unchanged. All statistics are fitted on training data, including sampled pre-contact anchors.

Each generated sequence uses one sampled contact map and one pose-noise vector shared across all requested motion times. The two branches use separate random streams. The contact-flow endpoint is clipped element-wise to $[0,1]$, while generated 6D rotations are orthonormalized before conversion to MANO poses. Both flows use 32 forward-Euler steps at inference.

\section{Additional qualitative comparisons}
\label{sec:supp_qual}
Figure~\ref{fig:supp_comparison} complements the eight-scene comparison in Figure~\ref{fig:comparison} with eight further scenes. It shows the approach, grasp, and post-grasp stages for DexGraspNet, LatentHOI, and our method. As in the main comparison, LatentHOI is a task-level baseline and does not use the depicted initial hand pose as input.

\end{document}